\documentclass[runningheads]{llncs}

\usepackage{eccv}

\usepackage{eccvabbrv}

\usepackage{graphicx}
\usepackage{booktabs}
\usepackage{wrapfig}

\newcommand{\UNetCN}{\mbox{UNet-CN }}

\newcommand{\UNetCNCP}{\mbox{UNet-CN-CP}}

\usepackage{amssymb}
\usepackage{hyperref}

\usepackage{orcidlink}

\begin{document}


\title{Performance-Carbon Trade-Offs across \\ Architectural Biases in Shear Flow Forecasting} 


\titlerunning{Performance-Carbon Trade-Offs across Architectural Biases}

\author{Sophia N. Wilson\inst{1}\orcidlink{0000-0002-4960-8223} \and
Jens Hesselbjerg Christensen\inst{2}\orcidlink{0000-0002-9908-8203} \and
Raghavendra Selvan\inst{1}\orcidlink{0000-0003-4302-0207}}

\authorrunning{S. N.~Wilson et al.}

\institute{Department of Computer Science, University of Copenhagen, Denmark
\email{\{sophia.wilson,raghav\}@di.ku.dk}\\
\and
Niels Bohr Institute, University of Copenhagen, Denmark\\
\email{hesselbjerg@nbi.ku.dk}}

\maketitle

\begin{abstract}
Development of modern deep learning methods has been driven primarily by the push for improving model efficacy (accuracy metrics), leading to large-scale models that require massive computational resources and result in considerable carbon footprint across the model lifecycle. In this work, we explore how architectural biases, specifically a model's receptive field and periodicity assumption, are associated with the trade-offs between predictive performance and carbon footprint for spatio-temporal forecasting of incompressible shear flow. We study seven models differing in these properties and evaluate pointwise accuracy, physics-fidelity, and carbon cost across training and inference.  
We find that no single model dominates across both predictive performance and carbon cost, that a lower training cost does not straightforwardly extend to inference, and that runtime alone is an unreliable proxy for carbon cost, particularly during training. Together, these results underscore the importance of explicitly evaluating carbon costs across the full model lifecycle.
We argue that model efficiency, alongside efficacy, should be a core consideration in machine learning model development and deployment.
\looseness=-1
\end{abstract}

\section{Introduction} \label{sec:intro}
Machine learning (ML) has advanced rapidly over the past decade, driven largely by the pursuit of improving model efficacy (accuracy metrics). This narrow focus has led to increasingly large models that require substantial computational resources and generate significant carbon emissions across their lifecycle~\cite{strubell_energyconsiderations_2019,anthony_carbontracker_2020,Sevilla_2022,luccioni2023estimating}. As models scale, efficacy gains diminish while costs escalate, underscoring the need to move beyond accuracy as the sole measure. A more holistic assessment must balance efficacy with efficiency, capturing not only predictive performance but also compute, energy, and carbon costs~\cite{henderson_2022}. This requires identifying strategies that strike a favourable balance rather than optimising for accuracy alone.

\begin{wrapfigure}{r}{0.45\textwidth}
    \centering
    \includegraphics[width=0.44\textwidth]{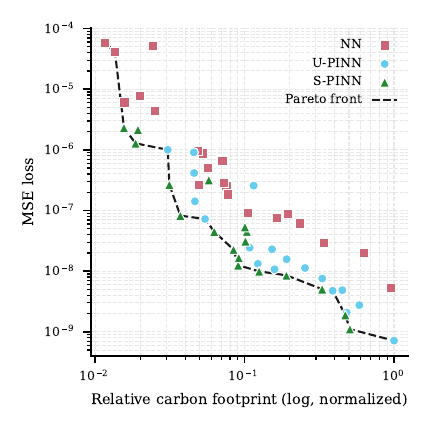}
    \caption{PINN toy experiment. Test mean squared error (MSE) versus CO$_2$eq emissions for a data-driven NN (pink squares), an unsupervised PINN (blue dots), and a semi-supervised PINN (green triangles). The dashed line marks the Pareto front, dominated by the S-PINN, indicating the best accuracy-carbon trade-off. 
    Data and code are adapted from~\cite{khoa_pinns_tutorial_2022}.}
    \vspace{-15pt}
    \label{fig:toy_example}
\end{wrapfigure}

Incorporating inductive biases into model design offers a promising path towards this balance. It is commonly understood in ML that strong model assumptions (bias) risk under-fitting, whereas high sensitivity to training data can lead to over-fitting (variance), and that striking the right balance is crucial to obtain the right class of models~\cite{kohavi1996bias}. We draw an analogy to model design under resource constraints: a stronger inductive bias can substitute for some of what would otherwise need to be learned from data, potentially reducing the data and compute required to reach a given level of performance. This suggests that inductive bias may trade off not only against variance, but also against the resource cost associated with developing a model.  

In this work, we consider inductive biases as an umbrella term spanning techniques such as Bayesian priors over model weights, task-specific regularisation terms, and architectural biases embedded in model design. 
Prior work has shown that embedding conservation laws, symmetries, or governing equations can reduce the number of trainable parameters~\cite{dutta_aq-pinns_2024, patra_PIKAN_2024}, reduce the amount of training data~\cite{Pestourie_2023, zhong_ECNN_2021, chen2021physics}, and accelerate training convergence~\cite{Brehmer_EquivarianceAtScale_2024, jahani_PINNs_2024}. 
Whether inductive biases can also reduce energy consumption and carbon footprint remains less explored. 

Fig.~\ref{fig:toy_example} offers a concrete illustration. In a simple experiment on a partial differential equation (PDE) system, we compare three models that share the same underlying architecture but differ in the strength of inductive bias imposed through their loss functions: a purely data-driven neural network (NN), an unsupervised physics-informed neural network (PINN) trained solely on PDE residuals, and a semi-supervised PINN combining data and PDE residuals. The semi-supervised PINN dominates the Pareto front, achieving lower prediction error at a lower carbon cost than either alternative. This suggests that striking the right balance between data and inductive bias may simultaneously improve accuracy and reduce the carbon footprint. 
\looseness=-1

Building on this intuition, we extend the perspective from task-specific models such as PINNs to architectures that embed inductive biases structurally. We study this in the context of spatio-temporal forecasting of PDE dynamics, 
specifically of a complex, high-resolution simulation of 2D incompressible shear flow. 
\looseness=-1

\newpage
Our key contributions towards this end are:
\begin{enumerate} 
    \item {\bf Carbon-aware evaluation:} We promote evaluation practices that explicitly account for training and inference carbon costs alongside predictive accuracy and physics-fidelity. 
    \vspace{10pt}
    \item {\bf Characterising the performance-carbon trade-off:} We provide an empirical characterisation of how architectural biases in receptive field and periodicity are associated with predictive performance and carbon footprint in spatio-temporal forecasting of incompressible shear flow. 
\end{enumerate}

\section{Background and Related Work}
\label{sec:background}


%

\paragraph{The Question of Resource-Awareness.} 
The standard metrics used to characterise the performance of ML models primarily focus on efficacy and are untethered from efficiency considerations. This introduces a preference for models that offer high efficacy at large resource costs~\cite{Bakhtiarifard_ECNAS_2024}. There are two primary streams of work that aim to include resource-aware metrics: composite metrics and multi-objective optimisation. 


Several composite metrics have been proposed to combine predictive performance with resources like energy consumption or carbon footprint. Evchenko \etal~\cite{evchenko2021frugal} introduced a composite metric that combines efficacy and efficiency metrics to study the influence of resource constraints on traditional (non deep learning) ML models. \textit{Carburacy}~\cite{moro2023carburacy} extends the idea to transformer models by jointly quantifying accuracy and carbon footprint. 
More recently, Kapoor \etal~\cite{kapoor2025beyond} proposed \textit{EcoL2}, a composite metric similar to \textit{Carburacy} but extended to include the costs of data generation and hyperparameter tuning.

Multi-objective optimisation has been primarily used in neural architecture search (NAS), resulting in models like EfficientNet~\cite{tan2019efficientnet}. More recently, energy consumption~\cite{Bakhtiarifard_ECNAS_2024} and carbon footprint~\cite{zhao2024end} have been taken into consideration in NAS. 
These methods, however, have not studied how architectural biases relate to the resource consumption of ML models.

\paragraph{Spatio-Temporal Forecasting Models and Architectural Biases.} There are many model classes for the task of spatio-temporal forecasting, differing in the architectural assumptions they encode. In this work, we focus on two: the UNet~\cite{ronneberger_2015} and the Fourier Neural Operator (FNO)~\cite{li2021fourier}, both originally developed for image-like data, as used in this study (Sec.~\ref{sec:experiment}). 
\looseness=-1

UNets perform forecasting by mapping a window of past states to future states, using convolutional layers that are inherently biased towards locality (each layer only mixes information within a limited receptive field) and translation equivariance (the same filters are applied everywhere in the domain). Skip connections fuse features across multiple spatial scales, but the model does not impose explicit constraints on the underlying dynamics beyond what is learned from data. By default, convolutions pad the input with zeros at the domain boundary, which implicitly assumes a non-periodic boundary, inconsistent with the periodic domain we study (Sec.~\ref{sec:experiment}). Circular padding corrects this inconsistency by replacing zero-padding with a periodic boundary assumption.
\looseness=-1

The FNO instead represents the state in Fourier space, letting the model mix information globally in a single layer rather than through stacked local convolutions. Like the UNet, it is translation-equivariant, since multiplication in Fourier space corresponds to a spatial convolution. In contrast to the UNet, however, its receptive field is global rather than local, and its use of the discrete Fourier transform assumes periodicity throughout the representation, rather than only at the domain boundary as with circular padding. Truncating to a finite number of Fourier modes additionally biases the model towards smooth, low-frequency solutions. We adapt the FNO from its original operator-learning formulation~\cite{li2021fourier} to an autoregressive setup.

\section{Experiment} 




\label{sec:experiment}
\paragraph{Problem Setup.} We consider spatio-temporal forecasting of dynamical systems, where the objective is to approximate the solution $u(x,y,t)$ of a PDE given initial and boundary conditions. We frame this as an autoregressive prediction on a two-dimensional, uniformly spaced mesh $(x,y) \in \{x_1,x_2,\ldots,x_H\} \times \{y_1,y_2,\ldots,y_W\}$ with discretised time steps $t \in \{t_1,t_2,\ldots,t_T\}$. At each step $t_k$ the state is represented as a tensor $u_k(x,y) \in \mathbb{R}^{H \times W \times C}$, where $k \in \{1,2,\ldots,T\}$ indexes discrete time steps in a trajectory of length $T$. $H$ and $W$ are spatial dimensions and $C$ is the number of physical fields (channels), such as pressure or velocity components. 
\looseness=-1

To capture temporal dependencies, the model is conditioned on a finite history of length $h$. The input window is denoted as $u_{k-h+1:k}(x,y) := (u_{k-h+1}(x,y), \\ \ldots, u_k(x.y))$. A parameterised function $f_\theta$ is trained to predict:
\begin{equation}
    u_{k+1}(x,y) = f_\theta \big(u_{k-h+1:k}(x,y)\big).
    \label{eq:autoregressive}
\end{equation}
During training, the loss is computed on one-step predictions. At inference, we assess long-term stability via autoregressive rollouts, obtained by iteratively applying Eq.~\ref{eq:autoregressive} and feeding the model’s own predictions back as inputs.

We instantiate this framework for forecasting incompressible shear flow governed by the Navier--Stokes equations, a setting that is both computationally demanding and governed by well-understood physical structure, making it well-suited for studying the efficacy-efficiency trade-off.
\looseness=-1

\begin{figure}[t]
    \centering
    \includegraphics[width=0.95\linewidth]{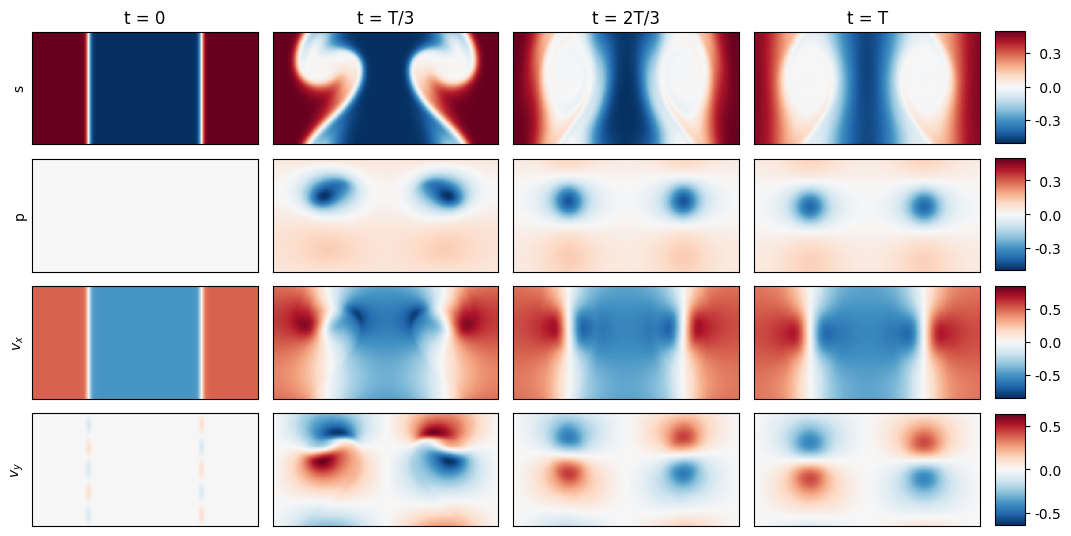}
    \caption{Example of the incompressible shear flow data. Temporal evolution of tracer ($s$), pressure ($p$), horizontal velocity ($v_x$) and vertical velocity ($v_y$) fields at four time steps: $t \in \{0, T/3, 2T/3, T\}$. The snapshots illustrate the coupled dynamics of the fields.}
    \label{fig:data}
\end{figure}

\paragraph{Data.} 
We use the incompressible shear flow data from \textit{The Well} collection~\cite{ohana_thewell_2024}, generated with \textit{Dedalus}~\cite{Burns_ShearFlow_2020}. 
The data describes 2D incompressible Navier--Stokes dynamics with an additional passive tracer,
\begin{align}
    \partial_t \mathbf{u} + \nabla p - \nu \nabla^2 \mathbf{u} &= -\mathbf{u}\cdot \nabla \mathbf{u}, \\
    \partial_t s - D \nabla^2 s &= -\mathbf{u}\cdot \nabla s, \\
    \nabla \cdot \mathbf{u} &= 0,
\end{align}
where $\mathbf{u}=(u_x,u_y)$ are the velocity components, $p$ is pressure satisfying $\int p = 0$, $s$ is the passive tracer, $\nu$ is viscosity, and $D$ is diffusivity, related to Reynolds and Schmidt numbers as $\nu=1/\text{Re}$ and $D=\nu/\text{Sc}$. The dataset captures the coupled non-linear evolution of these four fields on a periodic spatial domain (see example in Fig.~\ref{fig:data}). The system is initialised with a horizontal shear flow perturbed by small sinusoidal variations in the vertical velocity, localised at the shear layers; the tracer is initialised to match the shear flow and pressure to zero. Initial conditions are governed by three parameters: the number of shear layers $n_{\text{shear}} \in \{2,4\}$, the number of perturbation oscillations $n_{\text{blobs}} \in \{2,3,4,5\}$, and a width parameter $w \in \{0.25, 0.5, 1.0, 2.0, 4.0\}$ controlling the sharpness of the shear transition and localisation of the perturbations. Each field is standardised as $\hat{x}=(x-\mu)/\sigma$ prior to training.

We work with a subset, which consists of $240$ simulations spanning six PDE parameter settings and $40$ initial conditions settings. Simulations are resolved at $256\times 512$ with $200$ time steps ($\Delta t=0.1$). The data is split by initial condition into $80$/$10$/$10$ for training, validation, and testing, yielding $37,632$ one-step samples for training, $4,704$ for one-step inference, and $24$ trajectories for autoregressive rollout initialised at $t=0$.

\paragraph{Models.} We evaluate seven models that vary in receptive field and periodicity assumption, summarised in Table~\ref{tab:models}.
We consider two convolutional architectures with a local receptive field and no periodicity assumption: UNet and a ConvNeXt variant (UNet-CN)~\cite{liu_convnet_2022}, where the latter improves hardware efficiency through modern convolutional design. 
We introduce an explicit periodicity assumption by applying circular padding to both of these architectures (UNet-CP, \UNetCNCP), while keeping their local receptive field unchanged.
The remaining three models instead combine a global receptive field with an implicit periodicity assumption, achieved through a Fourier representation: a baseline FNO~\cite{li2021fourier}, a Tucker-factorised variant (TFNO)~\cite{kossaifi_TFNO_2023} that improves parameter efficiency, and a U-shaped variant (UFNO\footnote[1]{We denote this as UFNO for consistency with our FNO/TFNO naming, but it is not to be confused with the distinct U-FNO architecture by Wen~\etal~\cite{wen2022ufno}.})~\cite{rahman_UNO_2023} designed to better capture multi-scale interactions. 
Model configurations follow prior work: UNets, FNO, and TFNO follow Ohana~\etal~\cite{ohana_thewell_2024}, while UFNO follows Rahman~\etal~\cite{rahman_UNO_2023}. All models are scaled to a comparable size: 17.5M parameters for UNet and UNet-CN, 18.6M for UNet-CP and UNet-CN-CP, and 19.0M, 19.3M, and 18.8M for FNO, TFNO, and UFNO, respectively. Table~\ref{tab:config} summarises the full model configurations.
\looseness=-1

\begin{table}[t]
    \centering
    \caption{Architectural biases. 
    }
    \vspace{-5pt}
    \begin{tabular*}{\textwidth}{l @{\extracolsep{\fill}} l @{\extracolsep{\fill}} l}
    \toprule
    \textbf{Model} & \textbf{Receptive field} & \textbf{Periodicity}  \\
    \midrule
    UNet, UNet-CN         & Local  & None (zero-padded)           \\
    UNet-CP, UNet-CN-CP   & Local  & Explicit (circular padding)  \\
    FNO, TFNO, UFNO       & Global & Implicit (Fourier basis)     \\
    \bottomrule
    \end{tabular*}
    \label{tab:models}
\end{table}

\begin{table}[t]
    \centering
    \caption{Model configurations.}
    \label{tab:config}
    \vspace{-5pt}
    \resizebox{\textwidth}{!}{%
    \begin{tabular}{lccccc}
    \toprule
    \textbf{Model} & \textbf{Filter size/modes} & \textbf{Init dim} & \textbf{Blocks/stage} & \textbf{Up/down} & \textbf{Bottleneck} \\
    \midrule
    UNet, UNet-CP          & $3$        & $48$  & $1$ & $4$ & $1$  \\
    UNet-CN, UNet-CN-CP    & $7$        & $42$  & $2$ & $4$ & $1$  \\
    FNO, TFNO              & $16$       & $128$ & $4$ & --  & --  \\
    UFNO                   & $8,12,16$  & $128$ & $1$ & $3$ & $1$ \\
    \bottomrule
    \end{tabular}%
    \vspace{-5pt}
    }
\end{table}

\paragraph{Training Procedure.} All models use a history of $h=4$ states and are trained for one-step prediction according to Eq.~\ref{eq:autoregressive}. Across all models, we use a batch size of 16 and AdamW optimiser~\cite{loshchilov2019decoupled} with $10^{-4}$ weight decay. We adopt the coarse-tuned learning rates reported in \textit{The Well}~\cite{ohana_thewell_2024} ($5\cdot10^{-4}$ for UNets and $10^{-3}$ for FNOs, respectively), which are tuned across several datasets including the one used here. A linear warm-up cosine scheduler is applied for the first three epochs and the loss is the mean squared error (MSE) averaged across fields and space. Training run on a single Nvidia A40 GPU ($\leq$ 24h) with early stopping (patience 6). 
\looseness=-1

\paragraph{Evaluation Metrics.}
We assess forecasting performance using four metrics: two accuracy metrics, variance-scaled root mean squared error (VRMSE) and Pearson correlation coefficient (Pearson $r$), and two physics-fidelity metrics, velocity divergence and spectral energy error. 
VRMSE and Pearson $r$ are computed on all four physical fields (tracer, pressure, and both velocity components) and averaged across fields and trajectories. Velocity divergence and spectral energy error are computed only on the velocity field $\mathbf{u} = (u_x, u_y)$, since incompressibility and kinetic energy are properties of the velocity rather than of the tracer or pressure.

VRMSE reports the root mean squared error between a predicted field $\hat{y}$ and its ground truth $y$, normalised by the spatial variance of the ground truth. This normalisation accounts for the fact that the four fields differ substantially in their natural variability.
A VRMSE greater than one indicates that the prediction error exceeds the field's own fluctuations. 
Pearson $r$ quantifies the linear correlation between the prediction and ground truth. Unlike VRMSE, it is invariant to differences in scale and variance across fields, so a high $r$ can coexist with a systematically biased or miscalibrated prediction, which motivates the two physics-fidelity metrics below.

Velocity divergence quantifies violation of the incompressibility constraint (Eq.~4) in the predicted velocity field. We compute the divergence $\nabla \cdot \mathbf{u}$ via periodic central differences and summarise it at each rollout step as the spatial root mean square, averaged over trajectories,
\begin{equation}
    \text{Div}(\mathbf{u}) = \left\langle \left( \langle (\nabla \cdot \mathbf{u})^2 \rangle \right)^{1/2} \right\rangle_{\text{trajectories}},
\end{equation}
where $\langle \cdot \rangle$ denotes spatial averaging. 
Spectral energy error measures whether a model preserves the correct distribution of kinetic energy across spatial scales. We compute the radially-binned kinetic energy spectrum $E(\mathbf{u}; k)$ of the velocity field via its 2D discrete Fourier transform, binning by wavenumber $k = \sqrt{k_x^2+k_y^2}$ into $N_{\text{bins}} = \lfloor \min(H,W)/2 \rfloor$ shells, and report the relative $\ell_2$ error between the predicted and reference spectra, averaged over trajectories,
\begin{equation}
    \text{SpecErr}(\hat{\mathbf{u}}, \mathbf{u}) = \left\langle \frac{ \| E(\hat{\mathbf{u}};\cdot) - E(\mathbf{u};\cdot) \|_2 }{ \| E(\mathbf{u};\cdot) \|_2 + \epsilon } \right\rangle_{\text{trajectories}}.
\end{equation}

\paragraph{Carbon Footprint.} To estimate energy consumption during training and inference, we employ \textit{Carbontracker}~\cite{anthony_carbontracker_2020}. We report results in terms of carbon footprint rather than energy consumption directly, since energy is only problematic to the extent that it is generated from carbon-emitting sources. Were the grid entirely renewable, the same energy expenditure would carry negligible direct carbon emissions. In practice, however, the global energy mix remains far from carbon-neutral, and we therefore convert energy to carbon dioxide equivalent (CO$_2$eq) emissions using the global average carbon intensity for 2024 at $445$~gCO$_2$/kWh~\cite{iea_2025}. Using a fixed carbon intensity ensures fair comparisons across models, independent of when or where they were trained.

\section{Results}

\begin{figure}[t]
    \centering
    \includegraphics[width=\linewidth]{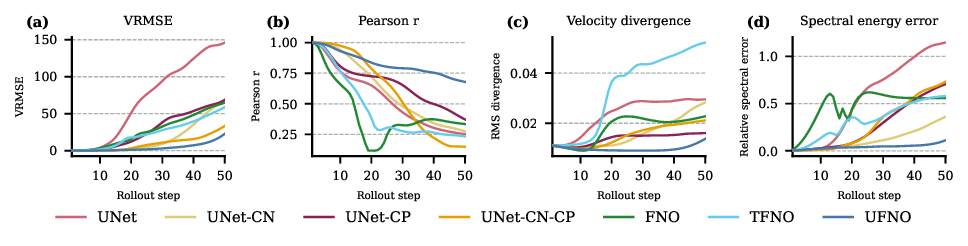}
    \caption{Rollout performance over 50 steps, averaged over the 24 test trajectories. {\bf a:}~Variance-scaled root mean squared error (VRMSE) and {\bf b:} Pearson correlation coefficient (Pearson $r$), each averaged over all four physical fields. {\bf c:} Velocity divergence and {\bf d:} spectral energy error, each computed on the velocity field only.}
    \label{fig:rollout_metrics}
\end{figure}
\begin{figure}[t]
    \centering
    \includegraphics[width=0.95\linewidth]{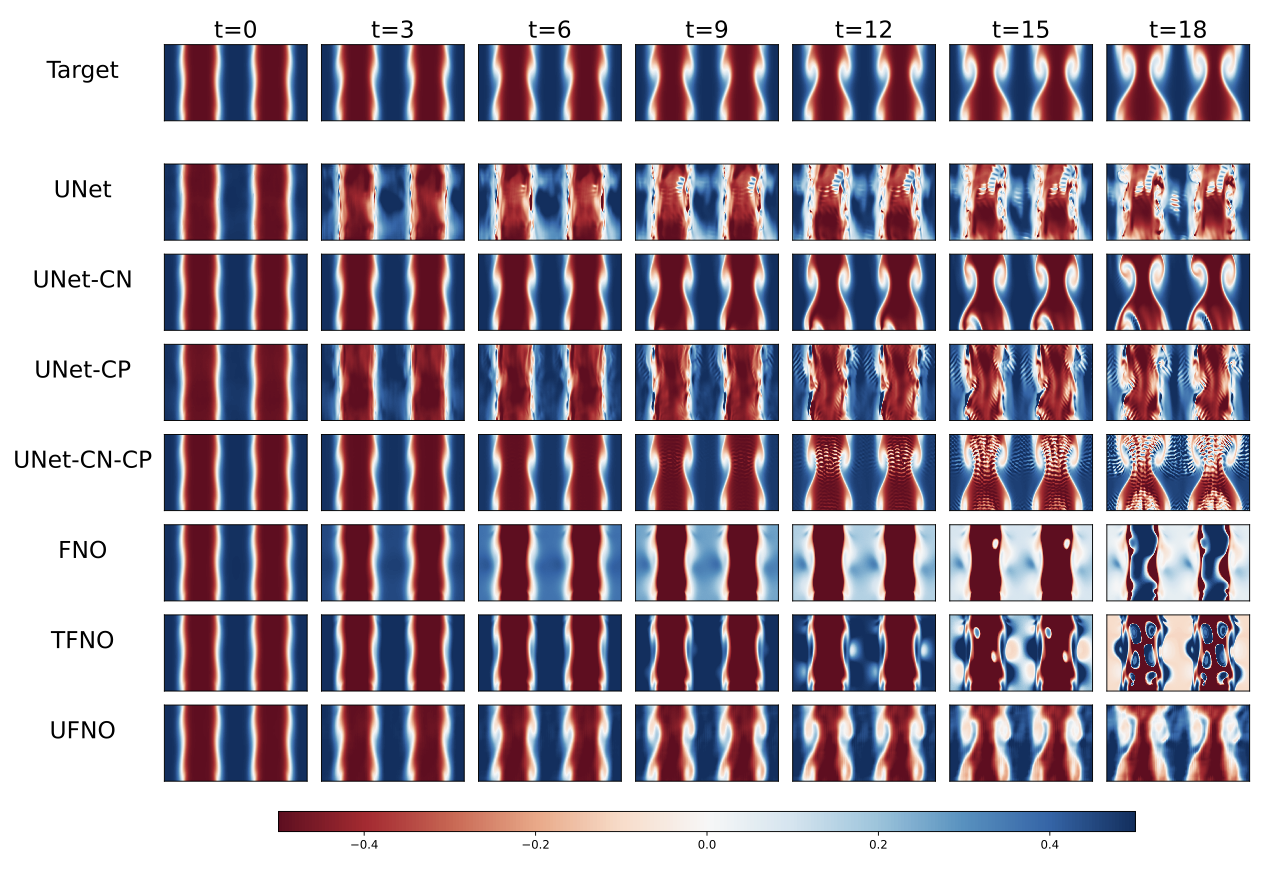}
    \caption{Example rollout predictions for the tracer field. Target trajectory (top row) compared with model predictions shown every third step of a 20-step rollout.}
    \vspace{-10pt}
    \label{fig:rollout_tracer}
\end{figure}

\paragraph{Predictive Performance.}
We first assess predictive performance at each step of a 50-step rollout across the four evaluation metrics described in Sec.~\ref{sec:experiment}.
As seen in Fig.~\ref{fig:rollout_metrics}, all models perform well for the first few steps, after which performance declines at different rates. 
Among the unpadded convolutional models, UNet-CN outperforms UNet by a wide margin across all evaluation metrics, showing that the modern convolutional design substantially improves both accuracy and physical fidelity.
Adding circular padding improves all four metrics for UNet, and improves or leaves roughly unchanged three of the four metrics for \UNetCN (with spectral energy error being the exception). This suggests that encoding the domain's periodic boundary explicitly tends to help, though the benefit is not uniform across architectures and evaluation metrics.
Among the neural operators, FNO and TFNO perform relatively poorly on all four metrics. In contrast, UFNO performs best across all four metrics, combining strong accuracy with low divergence and low spectral energy error. This suggests that the multi-scale U-shaped structure of UFNO, rather than the global receptive field alone, is what drives strong performance on this task.
An example of a visual rollout of the tracer field for all models can be seen in Fig.~\ref{fig:rollout_tracer}. 

\begin{table}[t]
\centering
\caption{Carbon footprint during training and inference. CO\textsubscript{2}eq emissions and run time for all models, measured with \textit{Carbontracker}~\cite{anthony_carbontracker_2020}. Inference results correspond to one-step forecasts over the full test set (4,704 samples) and 50-step rollouts over all test trajectories (24 samples). Inference costs are excluding the model-loading cost and are reported as mean plus/minus standard deviation over three runs. Inference times are reported as the mean over three runs.}
\label{tab:co2eq}
\resizebox{\textwidth}{!}{%
\begin{tabular}{lcccccccc}
\toprule
& \multicolumn{4}{c}{\textbf{Training}} & \multicolumn{2}{c}{\textbf{Inference (one-step)}} & \multicolumn{2}{c}{\textbf{Inference (rollout)}} \\
\cmidrule(lr){2-5} \cmidrule(lr){6-7} \cmidrule(lr){8-9}
\textbf{Model} & \textbf{CO\textsubscript{2}eq/epoch (g)} & \textbf{CO\textsubscript{2}eq total (kg)} & \textbf{\#Epochs} & \textbf{Time (h)} & \textbf{CO\textsubscript{2}eq (g)} & \textbf{Time (s)} & \textbf{CO\textsubscript{2}eq (g)} & \textbf{Time (s)} \\
\midrule
UNet       & 63  & 2.72 & 43 & 12.7 & 3.59 $\pm$ 0.03 & 58  & 2.11 $\pm$ 0.02 & 44 \\
UNet-CN    & 174 & 4.71 & 27 & 15.9 & 7.70 $\pm$ 0.05 & 113 & 2.66 $\pm$ 0.01 & 47 \\
UNet-CP    & 109 & 3.81 & 35 & 14.2 & 3.99 $\pm$ 0.05 & 62  & 2.15 $\pm$ 0.00 & 44 \\
UNet-CN-CP & 131 & 5.89 & 45 & 20.9 & 4.80 $\pm$ 0.02 & 73  & 2.24 $\pm$ 0.04 & 46 \\
FNO        & 67  & 0.80 & 12 & 4.0  & 7.80 $\pm$ 0.94 & 122 & 2.66 $\pm$ 0.29 & 50 \\
TFNO       & 68  & 1.09 & 16 & 5.3  & 7.14 $\pm$ 0.12 & 107 & 2.81 $\pm$ 0.21 & 52 \\
UFNO       & 37  & 1.12 & 30 & 3.6  & 8.94 $\pm$ 0.88 & 143 & 2.79 $\pm$ 0.05 & 48 \\
\bottomrule
\end{tabular}%
}
\end{table}

\begin{figure}[t]
    \centering
    \includegraphics[width=\linewidth]{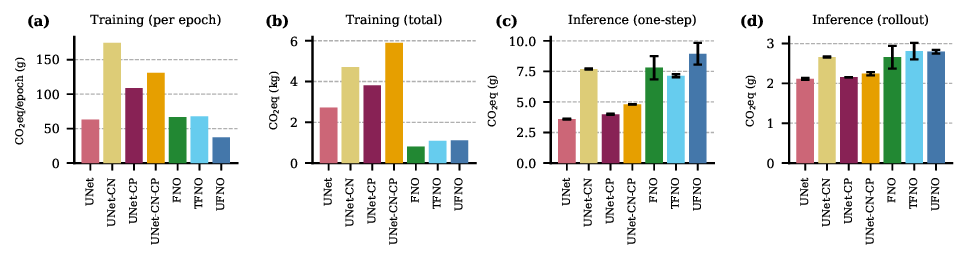}
    \caption{Carbon footprint of training and inference. {\bf a:} Training emissions per epoch (gCO$_2$eq). {\bf b:} Total training emissions (kgCO$_2$eq). {\bf c:} One-step inference emissions (gCO$_2$eq), summed over the full test set of 4,704 samples. {\bf d:} Rollout inference emissions (gCO$_2$eq), summed over 24 autoregressive rollouts of 50 steps. Training emissions are from a single run per model, while inference emissions are from three repeated runs excluding model-loading cost.}
    \label{fig:carbon}
    \vspace{-15pt}
\end{figure}

\paragraph{Carbon Footprint.}
We next evaluate how carbon footprint varies with receptive field and periodicity assumption across training and inference (see Table~\ref{tab:co2eq} and Fig.~\ref{fig:carbon}). 
Among the unpadded convolutional models, per-epoch training cost differs sharply: UNet is among the cheapest models to train (63~gCO$_2$eq/epoch), while UNet-CN is the most carbon-intensive of all seven (174~gCO$_2$eq/epoch, Fig.~\ref{fig:carbon}a). This gap carries over to inference, where UNet remains the cheapest model at both one-step (3.59~gCO$_2$eq) and rollout (2.11~gCO$_2$eq) inference, while UNet-CN is among the costliest at one-step (7.70~gCO$_2$eq) and moderately expensive during rollout (2.66~gCO$_2$eq). 

Adding circular padding affects the two backbones differently. For UNet, it raises per-epoch training cost substantially (63 to 109~gCO$_2$eq/epoch) while leaving inference cost almost unchanged (3.59 to 3.99~gCO$_2$eq one-step; 2.11 to 2.15~gCO$_2$eq rollout). For UNet-CN, adding circular padding instead lowers both carbon footprints, from 174 to 131~gCO$_2$eq/epoch during training, from 7.70 to 4.80~gCO$_2$eq at one-step inference, and from 2.66 to 2.24~gCO$_2$eq during rollout. The direction of the effect therefore depends on the backbone it is added to, rather than being a consistent property of the circular padding itself.

Among the neural operator variants, training and inference costs move in opposite directions compared to the convolutional models. FNO and TFNO train near to UNet's per-epoch cost (67~gCO$_2$eq/epoch and 68~gCO$_2$eq/epoch vs.\ 63~gCO$_2$eq/epoch), while UFNO trains at $55\%$ of that cost (37~gCO$_2$eq/epoch), the lowest of all. At inference, however, all three FNO variants are among the most expensive: UFNO has the highest one-step cost of all seven models (8.94~gCO$_2$eq) despite having the lowest training cost, and UFNO together with TFNO has the highest rollout cost (2.79~gCO$_2$eq and 2.81~gCO$_2$eq). 

\begin{figure}[t]
    \centering
    \includegraphics[width=\linewidth]{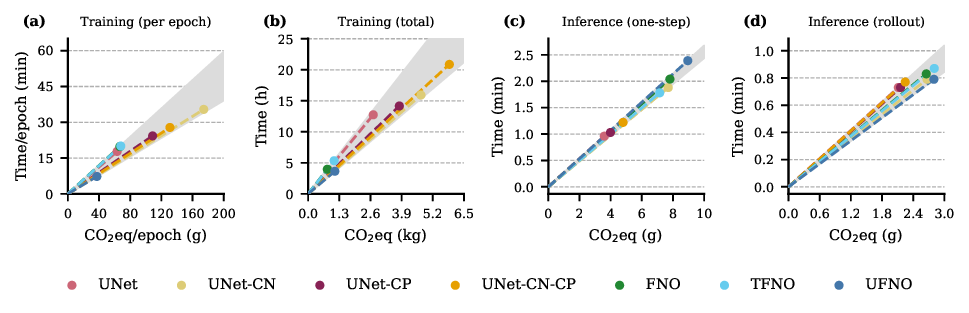}
    \caption{Runtime vs. carbon footprint. {\bf a:} Per-epoch training time vs. per-epoch training emissions (gCO$_2$eq). {\bf b:} Total training time vs. total training emissions (kgCO$_2$eq). {\bf c:}~One-step inference time vs. one-step inference emissions (gCO$_2$eq). {\bf d:} Rollout inference time vs. rollout inference emissions (gCO$_2$eq). For inference, each point marks a model's mean time and mean carbon footprint. The shaded region spans the ratios between the top and bottom dashed lines shown.}
    \label{fig:time_and_carbon}
\end{figure}


UFNO is therefore a clear example of a model that is cheap to train but expensive to deploy, underscoring the need to consider resource costs across the full lifecycle rather than a single stage.
A related pitfall is treating runtime as a proxy for energy consumption (or carbon footprint). Fig.~\ref{fig:time_and_carbon} shows that runtime and carbon footprint are correlated but not strictly proportional across models for both training and inference stages. This is due to different hardware operations having distinct power profiles, meaning that a model that executes faster does not necessarily have a lower carbon cost~\cite{henderson_2022}.

\section{Discussion}

Our study explores how architectural biases, specifically a model's receptive field and periodicity assumption, are associated with the trade-off between efficacy and efficiency in spatio-temporal forecasting of incompressible shear flow. The answer is not straightforward as carbon trade-offs vary across the model lifecycle. However, this motivates the need to consider resource costs across the entire model lifecycle. 
Introducing an explicit periodicity assumption via circular padding is associated with better performance, but its effect on training carbon efficiency and one-step inference is itself model-dependent, increasing both for UNet's padded counterpart (UNet-CP) and decreasing both for UNet-CN's padded counterpart (UNet-CN-CP).
The neural operators (FNO, TFNO, UFNO) achieve the lowest total training footprints among all seven models, but this comes at the expense of higher inference costs relative to the convolutional models. For example, UFNO's one-step inference cost is nearly twice that of the best-performing convolutional model (UNet-CN-CP), despite UNet-CN-CP having a training cost per epoch more than three times that of UFNO.
The lower total training cost of the neural operators is partly driven by fewer epochs to convergence under early stopping. As we use learning rates inherited from prior work rather than tuned per model, we cannot rule out that this untuned difference, rather than the architectural bias itself, drives the difference in convergence~speed.

In terms of predictive performance, UFNO is the only model that performs well across both pointwise accuracy and physics-fidelity metrics for most of the 50-step rollout. Other models, such as UNet-CN, UNet-CP, and UNet-CN-CP, also do reasonably well for the first 10--15 steps. Since the models were only trained on one-step predictions and not on longer rollouts, it is, however, expected that they will eventually accumulate errors. The model that is most attractive on accuracy and physical fidelity, UFNO, also has the lowest training cost per epoch, though not the lowest inference cost. This means that no single model dominates efficacy, training cost, and inference cost simultaneously.

Our results further show that runtime is not a fully reliable proxy for carbon cost at either stage, though the gap is significantly larger for training than for both one-step and rollout inference. Taken together, these results show that model selection cannot be based on efficacy alone, nor on carbon cost at a single stage, nor on runtime as a stand-in for energy or carbon cost. Amortising development costs across the model lifecycle is one way to factor in the carbon costs~\cite{cottier2024rising}.
\looseness=-1

Our results are specific to this PDE system, dataset subset, model family, and hardware setup, and should not be read as a general claim that architectural bias reduces carbon cost. We nonetheless believe that the broader question around incorporating inductive biases as a way to decrease data and compute requirements is worth raising since many scientific and engineering problems are governed by well-understood structures, and overlooking them is wasteful. 
However, inductive biases should not be incorporated at any cost, but only where they provides clear benefits in terms of predictive performance or resource efficiency. While our work centres on architectural biases, the same logic extends to other forms of prior knowledge: any regularising bias that constrains the learning task appropriately can potentially offer trade-offs between efficacy and efficiency.

\enlargethispage{2\baselineskip}
Ultimately, our study argues for a shift in mindset. Progress in ML should not be equated with ever-larger datasets, models, and compute budgets, but rather evaluated in terms of the value gained relative to the cost incurred~\cite{wilson2026hyper}. Learning everything from data while ignoring domain knowledge reflects an abundance mindset, in which resources are treated as unlimited. A more sustainable alternative is to exploit what we already know by embedding this knowledge into models. This work offers a step towards understanding how architectural biases relate to the efficacy-efficiency trade-offs, and argues for evaluating these explicitly. 
\looseness=-1

\paragraph{Limitations.}
Model configurations, including learning rates, follow prior work tuned on several PDE datasets, including the shear-flow dataset used here. Notably, the learning rate differs by a factor of two between the UNet and FNO families ($5\times10^{-4}$ vs.\ $10^{-3}$). This is a plausible confound for total training emissions, since these depend on the number of epochs to early stopping, which varies by roughly a factor of four across models (12--45). Training carbon and epoch counts are also reported from a single run per model, so we cannot rule out that run-to-run variance in convergence contributes to the reported differences.
We restricted our study to one PDE system, two model families, and a single GPU, and do not claim these findings generalise beyond this setting.
\looseness=-1

\section{Conclusions}

The primary message in this work, backed by empirical evidence, is to increase resource-awareness when designing ML experiments. While this can be accomplished in several ways~\cite{bartoldson2023compute}, we focus on the role of inductive biases and illustrate the different decisions practitioners have to consider. Using spatio-temporal forecasting of incompressible shear flow, we show how architectural biases are associated with the trade-off between predictive performance and carbon footprint.

This work advances a perspective grounded in resource-awareness and calls for a shift in mindset. As a community, we should discourage the pursuit of marginal improvements in performance that come from disproportionately higher resource costs. With our work, we have made the case to encourage resource-awareness by utilising inductive biases that encode existing domain knowledge.

\section*{Acknowledgements}
SW and RS acknowledge funding received from Independent Research Fund Denmark (DFF) under grant agreement number 4307-00143B. JHC acknowledges funding by DFF through the project Arctic Push, grant number 4258-00025B. RS also acknowledges funding received under European Union’s Horizon Europe Research and Innovation Action programme under grant agreements No. 101070284, No. 101070408 and No. 101189771. 

The authors thank Erik B Dam, Pedram Bakhtiarifard, Yijun Bian and others from the Machine Learning Section at UCPH for constructive feedback on the manuscript. The authors thank members of \href{https://saintslab.github.io/}{SAINTS Lab} for useful discussions throughout. 

%
%
\newpage
\bibliographystyle{splncs04}
\bibliography{main}


\end{document}